\documentclass[runningheads]{llncs}
\usepackage[T1]{fontenc}
\usepackage{graphicx}
\usepackage{booktabs}
\usepackage[misc]{ifsym}

\usepackage{hyperref}       
\usepackage{url}            
\usepackage{booktabs}       
\usepackage{amsfonts}       
\usepackage{nicefrac}       
\usepackage{microtype}      
\usepackage{xcolor}         
\usepackage{amsmath}
\usepackage{amsfonts}
\usepackage{dsfont}
\usepackage{amssymb}
\usepackage{subcaption} 
\usepackage{caption}
\usepackage{multirow}

\begin{document}
\title{Counterfactual Explanations for Time Series Should be Human-Centered and Temporally Coherent in Interventions}
\titlerunning{Time Series CFEs for Interventions}

%

\author{Emmanuel C. Chukwu \inst{1} \and
Rianne M. Schouten \inst{1} \and
Monique Tabak \inst{2} \and
Mykola Pechenizkiy \inst{1}}

\authorrunning{E.C. Chukwu et al.}

%
\institute{Data Mining Group, Eindhoven University of Technology,  the Netherlands
\and
Biomedical Signals and Systems Group, University of Twente, the Netherlands}

\maketitle

\begin{abstract}
Counterfactual explanations are increasingly proposed as interpretable mechanisms to achieve algorithmic recourse. However, current counterfactual techniques for time series classification are predominantly designed with static data assumptions and focus on generating minimal input perturbations to flip model predictions. This paper argues that such approaches are fundamentally insufficient in clinical recommendation settings, where interventions unfold over time and must be causally plausible and temporally coherent. We advocate for a shift towards counterfactuals that reflect sustained, goal-directed interventions aligned with clinical reasoning and patient-specific dynamics. We identify critical gaps in existing methods that limit their practical applicability, specifically, temporal blind spots and the lack of user-centered considerations in both method design and evaluation metrics. To support our position, we conduct a robustness analysis of several state-of-the-art methods for time series and show that the generated counterfactuals are highly sensitive to stochastic noise. This finding highlights their limited reliability in real-world clinical settings, where minor measurement variations are inevitable. We conclude by calling for methods and evaluation frameworks that go beyond mere prediction changes without considering feasibility or actionability. We emphasize the need for actionable, purpose-driven interventions that are feasible in real-world contexts for the users of such applications.

\end{abstract}
\section{Introduction}

In an ideal world, AI systems would be useful to all users in varying contexts and conditions. Yet, in many real-world applications, particularly in healthcare, users may seek not only to understand an unfavorable outcome but also to explore how they might achieve a more desirable one. For example, AI-powered digital health interventions have been shown to effectively support lifestyle changes in hypertension management \cite{leitner_effect_2024}, offering personalized guidance on activity, sleep, stress, and diet. Such systems illustrate the shift from static decision support to adaptive, goal-directed recommendations. This shift is central to the field of \textit{algorithmic recourse}, which focuses on producing explanations and actionable suggestions for individuals affected by automated decisions \cite{karimi2022survey}. A distinction arises between \textit{contrastive explanations} (CEs), which describe why an outcome occurred instead of an alternative, and \textit{consequential recommendations} (CRs), which suggest specific interventions that would alter the outcome. While both fall under the umbrella of counterfactual reasoning, CRs rely on stronger assumptions, such as causal invariance and stationarity: the idea that past actions would yield consistent results if repeated now \cite[footnote p.95:4]{karimi2022survey}. These assumptions raise practical concerns, especially in dynamic domains like healthcare.

A Counterfactual Explanation (CFE) is a post hoc interpretability method that identifies the minimal changes to input features required to produce the desired prediction \cite{wachter_counterfactual_2018}. CFEs hold the promise of explaining model behavior and supporting decision-making by answering contrastive (\textit{Why A, not B?}) and counterfactual (\textit{What if it had been B instead of A?}) queries \cite{stepin2021survey,keane_good_2020,verma_counterfactual_2024}. For instance, a CFE might show a physician what ECG signal pattern would result in a different diagnostic outcome or demonstrate to a developer how specific input modifications influence model confidence. However, CFEs are increasingly framed not just as explanations, but as recommendations. In e-health applications such as AI coaching systems \cite{leitner_effect_2024}, suggestions of increasing physical activity or adjusting the diet can be interpreted as counterfactuals to achieve better results. In this case, CFEs are expected to serve as both a model interpretability tool and a behavior-guiding recommendation mechanism. This extension demands additional constraints such as \textit{actionability}, \textit{plausibility}, \textit{sparsity}, and \textit{diversity}, which have led to the development of various evaluation metrics \cite{verma_counterfactual_2024,guidotti_counterfactual_2022,del_ser_generating_2024}.

We argue that the growing multiplicity of purposes, users, and evaluation metrics in CFE research creates ambiguity: it becomes unclear which methods are appropriate for which use cases. An effective CFE technique for system debugging may be misleading in a user-facing recommendation system. This ambiguity is particularly problematic in time series classification (TSC), where temporal dependencies make the generation of valid counterfactuals inherently more complex. Changing a past value can alter the plausibility of future values, and many counterfactual sequences, such as unrealistic changes to an ECG, may be infeasible for the user to act upon. These challenges require a careful re-evaluation of both the assumptions and evaluation criteria underpinning CFEs. This paper argues that CFEs for time series must reflect temporally coherent, causally plausible, and user-centered interventions rather than just input perturbations to change predictions. At a minimum, CFEs intended for recommendation should be robust to realistic variability, such as noise from human behavior or imperfect execution of suggested actions. We critically assess the current state of CFE evaluation, particularly in time series settings, and describe how new metrics and frameworks should better capture the demands of user-centered, real-world recommendations.

\section{Background}

Consider a fixed black-box classifier $f: \mathcal{X} \to \mathcal{Y}$ that takes an observation $\mathbf{x} \in \mathcal{X}$ as input and returns a probability vector $[0,1]^k$ for $k$ discrete output labels. We write $f_c(\mathbf{x})$ to denote the probability for class $c \in [1,k]$. The class label of the highest score is simply written as $\hat{y}$; the true label is $y$. In the tabular data case, $\mathbf{x} = (x_1, x_2, ..., x_d)$ is a d-dimensional vector taking values in 
$\mathcal{X}_1 \times ... \times \mathcal{X}_d$. Alternatively, we may consider $\mathbf{x} = (\mathbf{x}_1, \mathbf{x}_2, ..., \mathbf{x}_d) \in \mathbb{R}^{m \times d}$ to be a univariate ($d=1$) or multivariate $(d>1)$ time series, where $\mathbf{x}_j = (x^1_j, x^2_j, ..., x^m_j)$ is a sequence of values of length $m$ taken from the same feature space $\mathcal{X}_j$. The time in between successive measurements may be regular or irregular, may differ between features (channels), and could range from nanoseconds to several minutes, depending on the origin of the data. Obviously, incorporating temporal structure is of utmost importance in TSC tasks.

The field of algorithmic recourse is concerned with providing explanations and recommendations to individuals who are unfavorably impacted by the outcomes of black-box classifiers. Given an original input $\mathbf{x} \in \mathcal{X}$, a CFE is a modified input $\mathbf{x}'$ that changes the prediction from the current class $y$ to a different but desired class $y'$, with $y \neq y'$. Naturally, in generating CFEs, the model prediction $f(\mathbf{x}')$ should be valid and the modified input $\mathbf{x}'$ should be close in distance to the original instance $\mathbf{x}$. Following \cite{jiang_robust_2024}, a CFE is then computed as follows:

\begin{equation}\label{eq:CFE}
    \underset{\mathbf{x}' \in \mathcal{X}}{\arg\min} \; cost(\mathbf{x}, \mathbf{x}') \; \text{s.t.} f(\mathbf{x}) \neq f(\mathbf{x}'),
\end{equation}

\noindent where $cost(\mathbf{x}, \mathbf{x}') : \mathcal{X} \times \mathcal{X} \to \mathbb{R}^{+}$ is a distance metric in the input space. For differentiable classifiers, a constrained optimization approach (cf. \cite{wachter_counterfactual_2018}) can be used:

\begin{equation}\label{eq:CFEdiff}
    \underset{\mathbf{x}' \in \mathcal{X}}{\arg\min} \; loss(f(\mathbf{x}), f(\mathbf{x}')) + \lambda \cdot cost(\mathbf{x}, \mathbf{x}').
\end{equation}

\noindent Equation~(\ref{eq:CFEdiff}) expresses CFE generation as a differentiable loss, balancing two key properties: validity (left term) and proximity to the original input (right term). Section~\ref{sec:relatedwork} expands this by discussing additional properties relevant for the use of recommendations, such as actionability, feasibility, sparsity, diversity, and robustness, each with associated evaluation metrics.


Equations~(\ref{eq:CFE}) and~(\ref{eq:CFEdiff}) originate from the tabular setting, where instances $\mathbf{x} \in \mathbb{R}^d$. In TSC, Equation~(\ref{eq:CFE}) is often retained to define CFEs as modified inputs that change the predicted class. However, instead of using constrained optimization as in Equation~(\ref{eq:CFEdiff}), most TSC methods rely on pattern mining approaches, such as nearest unlike neighbors~\cite{delaney_instance-based_2021,ates_counterfactual_2021}, subsequence mining~\cite{bahri_discord-based_2024,bahri_shapelet-based_2022}, or genetic algorithms~\cite{hollig_tsevo_2022,refoyo_sub-space_2024}. However, cost metrics like $cost(\mathbf{x}, \mathbf{x}')$ are inherited from the tabular setting.

\subsection{Experimental setup}
We support our position with a small-scale evaluation of four representative CFE methods for time series classification (TSC): NG-CF~\cite{delaney_instance-based_2021}, CoMTE~\cite{ates_counterfactual_2021}, AB-CF~\cite{li_attention-based_2023}, and TSEvo~\cite{hollig_tsevo_2022} (summarized in Section~\ref{sec:relatedwork_tsc}). These methods were selected for their methodological diversity, citation impact, and prior rigorous evaluations. CFEs for NG-CF, CoMTE, and TSEvo were generated using the TSInterpret library~\cite{hollig_tsinterpret_2022}, with outputs verified against their original repositories. AB-CF was executed directly using its official implementation. All methods were run with default parameters, except for NG-CF (using the Nearest Unlike Neighbor (NUN) option) and TSEvo (limited to 200 epochs due to computational constraints).

We evaluate each method on four datasets from the UEA-UCR archive\footnote{\url{https://www.timeseriesclassification.com/}}, varying in dimensionality (univariate vs.\ multivariate), number of classes (2--5), and class distribution. Two datasets represent clinical signals: ECG and neurological activity. Table~\ref{tab:datasets} provides an overview of dataset characteristics. For each dataset-method pair, we randomly sampled $n=100$ test instances, generating one CFE per instance targeting a class different from the predicted label (the second-highest probability in multiclass cases). CFEs were considered invalid if they predicted the original class, and not all methods discovered a CFE for every instance. Table~\ref{tab:cfes_valid} reports the number of discovered, invalid, and valid CFEs. All subsequent analyses use the valid subset, with $n_{\text{valid}} \leq 100$.

\begin{table}[t!]
    \caption{Dataset characteristics.}
    \label{tab:datasets}
    \centering
    \begin{tabular}{lrrrcc}
    \toprule
    \textbf{Dataset} & \textbf{Training} & \textbf{Test} & \textbf{Length $m$} & \textbf{\#Classes $k$} & \textbf{\#Features $d$} \\
    \midrule
    ECG200 & 100 & 100 & 96 & 2 & 1 \\
    ECG5000 & 500 & 4500 & 140 & 5 & 1  \\
    TLECG & 23 & 1139 & 82 & 2 & 1 \\
    Epilepsy & 137 & 138 & 207 & 4 & 3 \\
    \bottomrule
    \end{tabular}    
\end{table}

For each test instance $\mathbf{x}_i$ and its counterfactual $\mathbf{x}'_i$, we evaluate the confidence of the predicted class and its validity, defined as $1$ if the predicted class matches the true (for $\mathbf{x}_i$) or target label (for $\mathbf{x}'_i$), and $0$ otherwise. To assess robustness, we introduce Gaussian noise to both $\mathbf{x}_i$ and $\mathbf{x}'_i$, and re-evaluate confidence and validity scores. This scenario, known as \textit{robustness against noisy execution (NE)}~\cite{jiang_robust_2024}, simulates deviations in input due to imprecise user action or noisy data acquisition. Specifically, we apply the perturbation function $\mathbf{x}_{\text{gauss}} = \mathbf{x} + \epsilon \cdot \mathcal{N}(0, \sigma_x)$, where $\sigma_x$ is the mean of the per-feature standard deviation of $\mathbf{x}$, and $\epsilon$ controls noise intensity. Noise is sampled to match the shape of $\mathbf{x}$ and added element-wise, approximating stochastic real-world variability. This aligns with recent formulations of robustness that consider both stochastic noise~\cite{pawelczyk_probabilistically_2023} and adversarial perturbations~\cite{dominguez-olmedo_adversarial_2022} as part of a broader robustness spectrum.

In our experiments, $\epsilon$ is varied from $0.0$ to $1.2$ to simulate increasing noise levels. We report the average confidence ($\textit{avgConf} \in [0,1]$) and validity ($\textit{avgVal} \in [0,1]$) across $n_{\text{valid}}$ test instances. This evaluation captures the model's resilience and the reliability of its counterfactual explanations under perturbations. The change in validity under noise reflects the \textit{invalidation rate} (IR)~\cite{pawelczyk_probabilistically_2023}, quantifying robustness as the average divergence between clean and noisy predictions. Metric definitions are provided in Section~\ref{app:exp_setup}. Code and results are available at \url{https://github.com/Healthpy/cfe_tsc_pos}.

\begin{table}[tb]
\centering
\caption{An overview of the number of invalid CFEs. 
Initial number of test instances ($n = 100$)
}
\label{tab:cfes_valid}
\begin{tabular}{lcccc}
\toprule
\textbf{Method} & \textbf{ECG200} & \textbf{ECG5000} & \textbf{Epilepsy} & \textbf{TLECG} \\
\midrule
AB-CF   & 40 & 37 & 9  & 0 \\
COMTE   & 10 & 5 & 3  & 0 \\
NG-CF   & 28 & 10 & -- & 0 \\
TSEVO   & 10 & 5 & 3  & 0 \\
\bottomrule
\end{tabular}
\end{table}

\section{Related work}\label{sec:relatedwork}



Interpretability and explainability are critical to building trust in machine learning, particularly in high-stakes domains like eHealth\cite{chou_counterfactuals_2022,ali_explainable_2023}. CFEs provide local interpretability by identifying how an input must change to alter a model’s prediction~\cite{keane_good_2020}. However, in this paper, we emphasize that explaining model behavior is not equivalent to providing actionable recommendations for recourse. Depending on the system and the user context, CFEs must satisfy more than just validity (producing the desired prediction) and proximity (similarity to the original input)~\cite{verma_counterfactual_2024}. Additional properties become essential, including sparsity (minimal changes in characteristics), diversity (generating multiple and varied CFEs), and plausibility (changes that remain realistic or adhere to the datamanifold)~\cite{verma_counterfactual_2024}. Critically, \emph{actionability}: whether users can interpret and implement the recommended changes must also be considered~\cite{guidotti_counterfactual_2022,ustun_actionable_2019}.

Furthermore, when using CFEs in real-world applications, it is crucial that they retain their validity under minor variations, such as small changes in the original input or slight deviations in the proposed change vector. Real-world data are often noisy and imprecise and users may not be able to precisely execute the recommended changes. In this context, Jiang et al. \cite{jiang_robust_2024} evaluate the CFE methods in four types of robustness. First, robustness to \emph{model changes} (MC) ensures that CFEs remain valid even when the model $f$ is retrained or slightly altered. Second, robustness to \emph{model multiplicity} (MM) addresses the variability between models that perform similarly but produce different predictions for the same input CFEs and should remain stable across this diversity. Third, robustness to \emph{noisy execution} (NE) reflects the imperfect implementation of the recommendations; small deviations should still produce the desired result. Fourth, robustness to \emph{input changes} (IC) requires similar inputs to produce similar CFEs, supporting fairness and interpretability when users have nearly identical profiles. An alternative categorization of \cite{guyomard_generating_2023} distinguishes the robustness to modifications in MC, IC, and CFE (related to NE in \cite{jiang_robust_2024}). To assess these properties, \cite{verma_counterfactual_2022} and \cite{jiang_robust_2024} list several evaluation metrics, developed mainly for tabular data. Proximity is commonly measured using $\ell_1$ and $\ell_2$ norms, while sparsity is quantified using the $\ell_0$ norm~\cite{kan_benchmarking_2024}. Validity and robustness are evaluated through changes in predicted probabilities $f_c(\mathbf{x})$, such as \emph{validity after retraining} (VaR) or \emph{validity after perturbation} (VaP)~\cite{jiang_robust_2024}. Additional techniques include adversarial perturbation analysis, noise-based sampling, and diversity evaluations~\cite{leofante_promoting_2023}. However, these metrics, rooted in static data, may not fully address practical challenges in the real world involving temporal data.

In the context of time series, evaluation strategies are generally similar to those for tabular data. Instead, in Section \ref{sec:temp}, we argue that these metrics may not be sufficient for evaluating CFE methods for TSC, since time series are inherently sequential and an input change in one value may not be achieved without changing the previous or the next value. To the best of our knowledge, no survey on CFE methods for TSC has addressed these implications, despite the growing recognition that contributions in this area must extend beyond algorithmic performance to real-world applicability. In this paper, we provide an overview of the existing TSC CFE methods in Table \ref{tab:cf_methods}. The methods vary greatly in whether they modify single time points \cite{hollig_tsevo_2022}, segments \cite{delaney_instance-based_2021,li_attention-based_2023}, or entire channels \cite{ates_counterfactual_2021}. 
Many methods build on pattern mining methodology such as motif discovery \cite{li_motif-guided_2024,bahri_discord-based_2024}, shapelets \cite{li_sg-cf_2022,bahri_shapelet-based_2022,huang_shapelet-based_2024,li_reliable_2024}, nearest unlike neighbors \cite{ates_counterfactual_2021,delaney_instance-based_2021} and saliency maps \cite{li_cels_2023,spinnato_understanding_2023}. Some methods rely on genetic algorithms \cite{hollig_tsevo_2022,refoyo_sub-space_2024}; others use gradient-based modifications of the input space \cite{wang_glacier_2024,wang_counterfactual_2021}. 

\begin{table}[b!]
\centering
\caption{Overview of Counterfactual Explanation Methods for Time Series Classification. We distinguish methods based on Distance metrics (DiOp), Shapelets (ShOp), Gradients (GrOp), Adversarial learning (AlOp), Evolutionary algorithms (EvOp), Reinforcement learning (RLOp), and Causality (Causal). U/M refers to applicability in univariate or multivariate time series.}
\label{tab:cf_methods}
\begin{tabular}{l|l|l|c|c}
\hline
\toprule
\textbf{Paper} & \textbf{Method} & \textbf{Mechanism} & \textbf{Category} & \textbf{U/M} \\ \hline
\cite{ates_counterfactual_2021} & CoMTE & Channel substitution via greedy search & DiOp & M \\ \hline
\cite{delaney_instance-based_2021} & NG-CF & Segment substitution based on NUNs & DiOp & U \\ \hline
\cite{li_sg-cf_2022} & SG-CF & Shapelet-guided transformation & ShOp & U, M \\ \hline
\cite{bahri_shapelet-based_2022} & SETS & Shapelet coefficient modification & ShOp & M \\ \hline
\cite{filali_boubrahimi_mining_2022} & TimeX & Barycenter averaging and saliency maps & DiOp & U, M \\ \hline
\cite{li_cels_2023} & CELS & Saliency map-guided perturbations & GrOp & U \\ \hline
\cite{li_m-cels_2024} & M-CELS & Saliency map-guided perturbations & GrOp & M \\ \hline
\cite{wang_glacier_2024} & Glacier & Gradient search in original or latent space & GrOp & U \\ \hline
\cite{huang_shapelet-based_2024} & Time-CF & Shapelet extraction and TimeGAN generation & ShOp, AlOp & U, M \\ \hline
\cite{wang_counterfactual_2021} & LatentCF++ & Latent space perturbation with autoencoders  & GrOp & U, M \\ \hline
\cite{bahri_discord-based_2024} & DiscoX & Matrix Profile discord replacement & DiOp & U, M \\ \hline
\cite{li_attention-based_2023} & AB-CF & Attention-based segment modification & DiOp & U, M \\ \hline
\cite{bahri_temporal_2022} & TeRCE & Temporal rule mining with shapelets & ShOp & M \\ \hline
\cite{van_looveren_conditional_2021} & \( \text{G}_\text{CF} \) & Conditional generative model &  AlOp & U, M \\ \hline
\cite{li_motif-guided_2024} & MG-CF & Motif-based subsequence replacement & ShOp & U, M \\ \hline
\cite{hollig_tsevo_2022} & TSEvo & Multi-objective evolutionary search & EvOp & U, M \\ \hline
\cite{sun_counterfactual_2024} & CFWoT & RL-based sequential decision-making & RLOp & M \\ \hline
\cite{lang_generating_2023} & SPARCE & GAN-based sparse counterfactuals & AlOp & M \\ \hline
\cite{spinnato_understanding_2023} & LASTS & Saliency maps, instance-based & Hybrid & U, M \\ \hline
\cite{yan_self-interpretable_2023} & CounTS & Variational Bayesian causal modeling & Causal & M \\ \hline
\cite{refoyo_sub-space_2024} & Sub-SpaCE & Genetic algorithms & EvOp & U \\ \hline
\bottomrule
\end{tabular}
\end{table}

\subsection{Detailed summary of 4 distinguished CFE methods for TSC}\label{sec:relatedwork_tsc}

In this paper, we experimentally evaluate the performance of 4 CFE generation methods for TSC. First, NG-CF\cite{delaney_instance-based_2021} is among one of the early strategies, including those based on K-Nearest Neighbors (KNN) or NUN, have inspired local perturbation methods such as Native Guides~\cite{delaney_instance-based_2021}. These methods are confined largely to univariate settings and are heavily reliant on simplistic heuristics like proximity and sparsity. It follows two steps: (i) retrieving the Native Guide, where the closest instance from a different class is selected, and (ii) adapting the Native Guide, where the instance is iteratively perturbed toward the query using distance metrics like dynamic time warping (DTW) or feature weight vectors (e.g., Class Activation Mapping) to modify key subsequences. 

Second, CoMTE~\cite{ates_counterfactual_2021} is a multivariate extension. 
Although efficient, such methods ignore broader temporal and semantic structures, producing counterfactuals that may appear valid locally but lack global interpretability. CoMTE produces counterfactuals in a multivariate context by perturbing the channels of a time series using a heuristic method. They modified the initial approach of Wachter et al\cite{wachter_counterfactual_2018}. by substituting the point-wise distance function \(d\) with one that operates on a channel-by-channel basis.

Third, AB-CF \cite{li_attention-based_2023} focuses on identifying and perturbing the most important segments of a time series using an attention mechanism. By narrowing attention to a small set of influential segments, AB-CF ensures that the generated counterfactuals are valid, sparse, and interpretable. The method first generates a pool of candidate subsequences and then selects one or more per time series to replace, based on their contribution to the model prediction, quantified using the Shannon entropy algorithm. This targeted modification strategy enables AB-CF to produce efficiently computed counterfactuals. Focusing on locality, the method struggles to represent overall temporal relationships, leading to counterfactuals valid locally but implausible globally.

Fourth, TSEvo\cite{hollig_tsevo_2022} extends the concept of CFEs to both univariate and multivariate time series classification. It formulates an optimization problem that balances three key objectives: proximity, sparsity, and output distance. This multiobjective optimization problem is solved using a genetic algorithm. The algorithm employs crossover and mutation operations, applied both at the level of individual time series values and larger segments, allowing it to explore counterfactuals tailored for time series data. Unfortunately, the method is computationally expensive.

The effectiveness of CFE methods ultimately depends on the utility of their CFEs from a human-centered perspective. We argue that beyond technical correctness, counterfactuals are valuable only to the extent that they are interpretable, actionable, and robust in practical decision-making contexts.

\section{CFEs for TSC have a temporal blind spot}\label{sec:temp}

It is obvious that in TSC tasks, the incorporation of temporal structure is of utmost importance, and valuable methods have been developed specifically for this task \cite{middlehurst_bake_2023}. However, we observe that such time awareness is lacking in methods generating CFEs for those same classification tasks. In addition, we observe that many existing evaluation metrics are borrowed from the tabular data case: These metrics do not account for temporality, which limits their real-world utility, particularly in longitudinal decision-making scenarios. 

Consider Diameter App \cite{hietbrink_digital_2023}, an e-health application that aims to support patients with Type 2 diabetes to understand blood glucose fluctuations and the relationship with certain lifestyle behavior choices \cite{schouten_mining_2022,den_braber_glucose_2021,danne_international_2017}. Here, it would be valuable to have CFEs suggesting a reduced carbohydrate intake at time $t$ to prevent hyperglycemia. Although such a CFE may be locally valid, the intervention can intentionally affect the glucose level at $t+1$. A CFE should be temporally aware and consider downstream effects rather than instantaneous classification shifts. It thus seems more appropriate to first change the input sequence at one time point, then model its effect on the rest of the input sequence, and then evaluate the potential change in prediction. 

A step in the right direction would be to consider modifying subsequences rather than changing the entire time series. Three of the four CFE methods that we consider in this paper take this approach (NG-CF \cite{delaney_instance-based_2021}, AB-CF \cite{li_attention-based_2023} and TSEvo \cite{hollig_tsevo_2022}). The fourth method, CoMTE \cite{ates_counterfactual_2021}, modifies entire channels (although it aims to reduce the number of channels). Nevertheless, it is questionable whether these kind of input changes could be considered feasible for end users like patients and individual persons: one could probably directly influence the blood glucose value at time $t$, but not again at time $t+1$, $t+2$, and so on, without taking into account the effect of earlier changes. In this regard, robustness becomes a critical property in time series CFEs, since minor perturbations at any point could affect consequent events and disturb the semantic integrity of the sequence and its classification accuracy. Sections \ref{sec:temp_demo} and \ref{sec:user_demo} provide demonstrations.

In addition, we argue that CFEs for TSC should not be evaluated solely with evaluation metrics developed for the tabular data case. Indeed, in addition to evaluating the validity and confidence of CFE predictions, the formal definition of a CFE as given in Equation (\ref{eq:CFE}) persuades us to evaluate their $cost(\mathbf{x},\mathbf{x}') \in \mathcal{X} \times \mathcal{X}$. Many existing CFE methods for TSC use common distance metrics such as the $\ell_1$ and $\ell_2$ norm to evaluate $cost(\mathbf{x},\mathbf{x}')$ as proximity. For multivariate time series data, these metrics have high dimensionality. In addition, they do not reflect other aspects that strongly relate to the concept of proximity, such as whether two sequences have a similar periodicity or amplitude. We contend that other evaluation metrics are needed, not just for evaluating proximity but also for evaluating other important CFE properties such as sparsity. In Section \ref{sec:temp_demo}, we demonstrate that small adjustments to existing metrics are insufficient.  


\subsection{Demonstration}\label{sec:temp_demo}



\begin{figure}[bt]
\centering
    \includegraphics[width=0.85\linewidth]{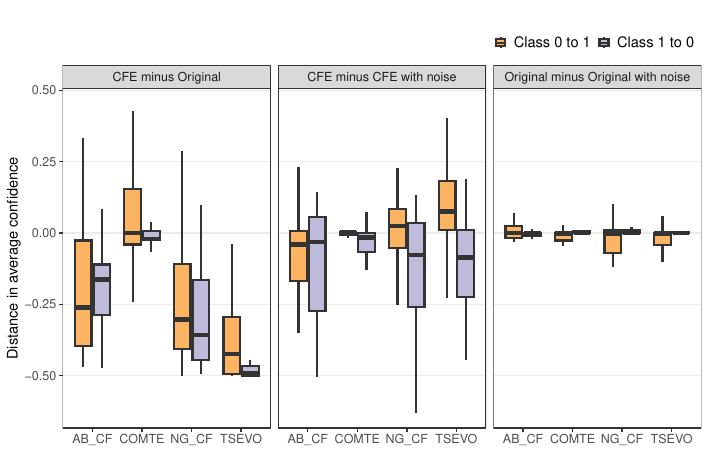}
    \caption{Confidence distance between original and CFE (left), CFE with/without $\epsilon=0.2$ Gaussian noise (center), and original input with/without noise (right), for $n_{\text{valid}}$ CFEs from 4 methods on ECG200. Results are shown separately for CFEs targeting class 1 (orange) and class 0 (purple).}

    \label{fig:ecg200-dif}
\end{figure}

Figure \ref{fig:ecg200-dif} compares the confidence of the predicted classes between CFEs and their original instances for the ECG200 dataset. Recall from Table \ref{tab:datasets} that ECG200 is a small dataset with $N=100$ training examples, $k=2$ classes, $d=1$ feature and a sequence length of $m=96$. Table \ref{tab:cfes_valid} displays that many CFEs were not generated at all (for the AB-CF and NG-CF methods), or considered invalid (for all 4 CFE methods). We now discuss results in Figure \ref{fig:ecg200-dif} (see Table \ref{tab:ecg200-dif} in Section \ref{sec:app_results} for detailed information) for the $n_{\text{valid}}$ instances only.

First, we find that the classification model is robust to Gaussian noise with $\epsilon=0.2$. This can be seen in the third, most right panel, where the difference in prediction probabilities between instances with and without noise are close to zero. Differences between CFE methods may occur due to randomness in the generated noise. In contrast to original input, adding Gaussian noise to CFEs notably affects the predicted class probability (center panel of Figure~\ref{fig:ecg200-dif}). For all four methods, prediction confidence generally decreases under noise, except in two cases. First, for NG-CF and TSEvo, confidence in predicting class 1 \textit{increases} after noise is added, as seen in the orange boxplots on the right of the center panel (values greater than 0). This indicates that CFE robustness may vary between groups of instances, possibly due to training-related issues such as class imbalance. Such disparities are problematic in practice, as they suggest that some users may receive more (or less) stable counterfactuals. Evaluation protocols should therefore go beyond global metrics and assess fairness between subgroups, as discussed in further detail in Section~\ref{sec:user}.

\begin{figure}[bt]
\centering
    \begin{subfigure}[b]{0.46\textwidth}
            \includegraphics[width=\textwidth]{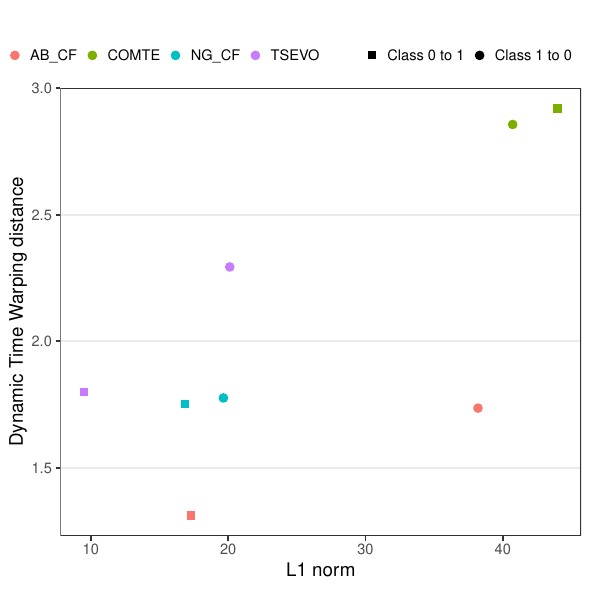}
            \caption{Proximity}
            \label{fig:ecg200-prox}
    \end{subfigure}\hfill
    \begin{subfigure}[b]{0.44\textwidth}
            \includegraphics[width=\textwidth]{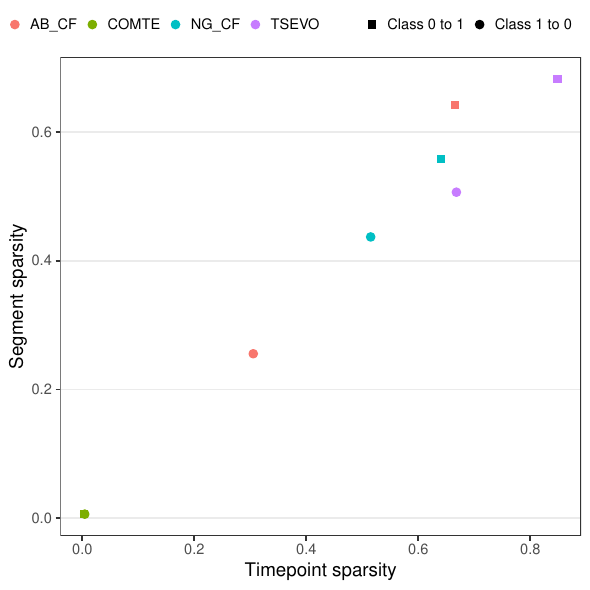}
            \caption{Sparsity}
            \label{fig:ecg200-spars}
    \end{subfigure}
    \caption{Comparison of the average proximity and average sparsity for 4 CFE methods on the ECG200 dataset. (a) Proximity is measured with $\ell_1$ norm (x-axis) and the DTW distance (y-axis). (b) Sparsity is measured by evaluating changes in time points (x-axis) and changes in entire segments (y-axis). Results are displayed separately for CFEs generated for class 1 (rather than 0, with square) and class 0 (rather than 1, with circle).}
    \label{fig:ecg200-proxspars}
\end{figure}

Second, CoMTE exhibits minimal change in prediction confidence when Gaussian noise ($\epsilon = 0.2$) is added, clearly shown in both the center and left panels of Figure~\ref{fig:ecg200-dif}. Unlike other methods, CoMTE’s CFEs retain similar confidence to the original inputs. This is likely because CoMTE relies on NUN substitution~\cite{ates_counterfactual_2021}, directly replacing the input with a near-instance from another class. Although this produces high-confidence counterfactuals by design, such CFEs may lack meaningful interpretability or personalization. Despite its robustness performance, the reliance of CoMTE on NUNs raises concerns. As shown by the green dots in Figure~\ref{fig:ecg200-proxspars}, these CFEs exhibit low sparsity and high proximity. We measure proximity (Figure~\ref{fig:ecg200-prox}) using both $\ell_1$ norm and Dynamic Time Warping (DTW), and sparsity (Figure~\ref{fig:ecg200-spars}) using two metrics: the traditional count of changed input values and an adapted metric counting modified segments. The results are consistent across both formulations and reveal a near-linear trade-off between proximity and sparsity. This shows that high scores on traditional robustness or proximity metrics do not necessarily imply useful or interpretable counterfactuals.

\subsection{Alternative Views}

Unlike tabular data, time series are inherently sequential; modifying one time step may disrupt temporal dependencies or produce unrealistic patterns, even if the perturbation is sparse. Consequently, standard pointwise proximity metrics are insufficient: evaluation must consider global shape, alignment, and temporal dynamics. While several works adapt existing metrics for time series \cite{delaney_instance-based_2021,ates_counterfactual_2021,bahri_temporal_2022,refoyo_sub-space_2024}, we argue that current CFE evaluation strategies for TSC are fragmented: too many metrics exist, and too few account for temporal structure meaningfully.

A growing body of work calls for a more standardized and holistic evaluation of explainability methods. Bhattacharya et al.~\cite{bhattacharya2024good} advocate a framework that spans model, data, prediction, and user dimensions, stressing that evaluation must move beyond isolated metrics. Artelt et al.~\cite{arteltEvaluatingRobCf2021} recommend prioritizing plausibility over proximity to improve fairness and robustness, while Nguyen et al.~\cite{NguyenRobustFramework2023} propose the AMEE framework, assessing the robustness of the explanation via perturbations between classifiers. Without such standardization, evaluation inconsistencies can lead to misleading conclusions and unreliable deployment~\cite{verma_counterfactual_2024,daza_causal_2018,runge_causal_2023}. Tools like TSInterpret~\cite{hollig_tsinterpret_2022} support multiple CFE methods (e.g., TSEvo~\cite{hollig_tsevo_2022}, NG-CF~\cite{delaney_instance-based_2021}, CoMTE~\cite{ates_counterfactual_2021}) and offer visualizations, while XTSC-Bench~\cite{hollig_xtsc-bench_2023} provides benchmarking across models and datasets. Kan et al.~\cite{kan_benchmarking_2024} further benchmark five TSC CFE methods, proposing new metrics tailored for time series. Yet, even with such tools, fixed evaluation sets, regardless of their temporal design, are insufficient on their own.

We argue that CFEs for TSCs must be developed and evaluated within the context of their intended application. For example, CFEs supporting ML debugging differ significantly from those offering behavioral recommendations to patients or treatment insights to clinicians. Time series inputs from wearables (e.g., ECG, HR, step counts) vary in modifiability: while physical activity can be altered, users cannot change their ECG. Thus, clinical feasibility cannot be assumed to be the result of algorithmic success. It is essential to ask: who is the user, what can they act upon, and in what context will CFEs be used? Without these considerations, even CFEs that score well on existing metrics may fail in practice.

\section{CFEs Should be User-Centered and Recourse-Aware} \label{sec:user}

Classic CFE algorithms are typically task-centered, focused on flipping model predictions with minimal input changes. However, CFEs are increasingly used not only for interpretation, but also as a foundation for recommendations, guiding users on how to achieve a desired outcome~\cite{verma_counterfactual_2024,guidotti_counterfactual_2022,karimi2021algorithmic}. This requires that CFEs align with user goals, prior knowledge, and domain-specific constraints. Effective recommendations must account for the user’s ability to interpret and implement changes. Without this user-centered perspective, CFEs may remain technically valid, yet practically irrelevant or misleading.

\subsection{Demonstration} \label{sec:user_demo}

\begin{figure}[tb]
\centering
    \includegraphics[width=0.85\linewidth]{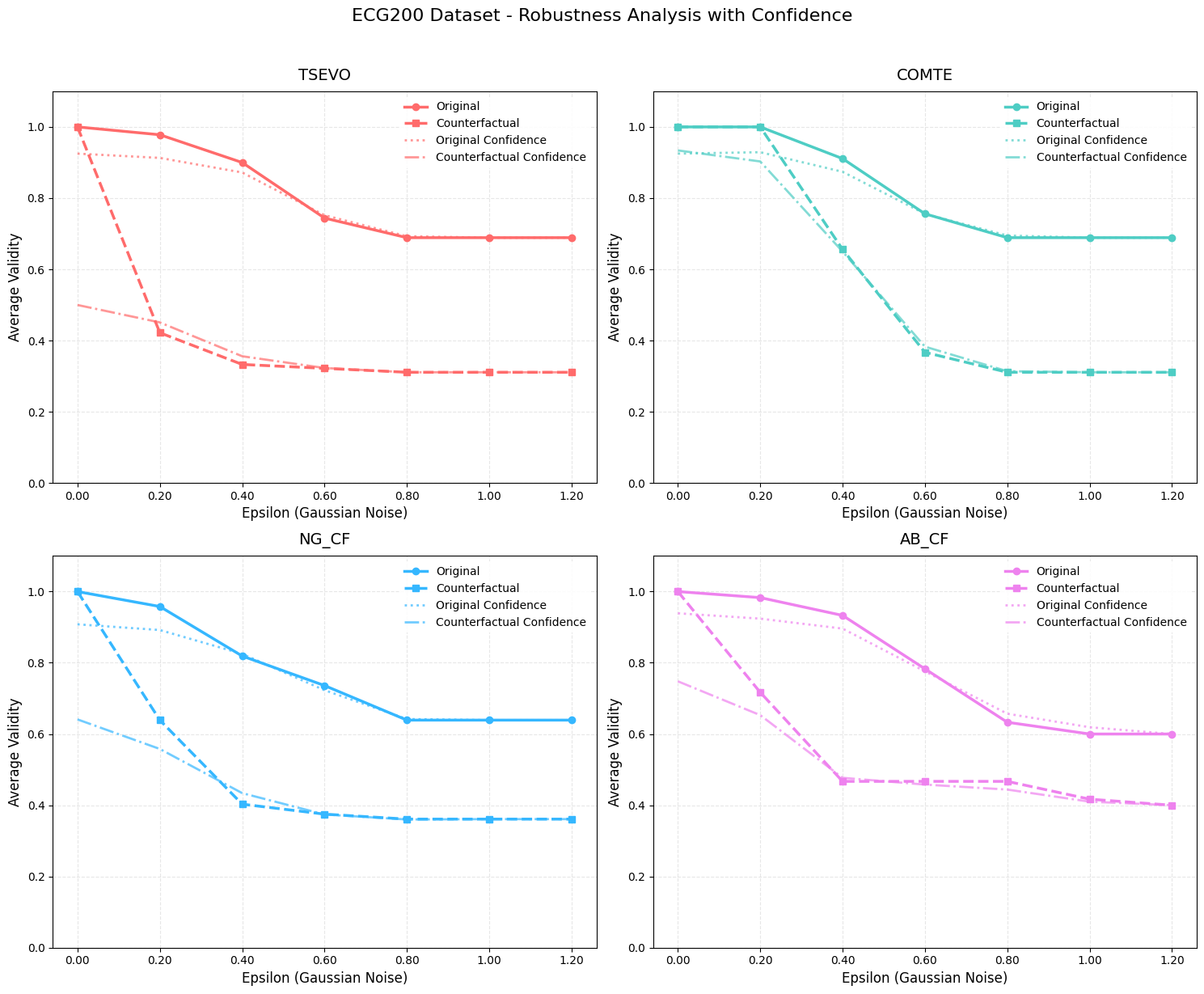}
    \caption{Robustness analysis on the ECG200 dataset showing average validity under increasing Gaussian noise for four counterfactual methods. Solid lines represent original instances; dashed lines represent counterfactuals. Confidence levels are shown with dotted and dash-dotted lines. Original examples consistently show greater robustness to noise than their counterfactual counterparts.}
    \label{fig:cf-methods-ecg200}
\end{figure}

We illustrate this issue using results from the ECG200 dataset. Additional results for ECG5000, Epilepsy and TwoLeadECG are provided in the Appendix~\ref{app:B}, with highlights in Section~\ref{sec:results_otherdatasets} only when deviations occur. Figure~\ref{fig:cf-methods-ecg200} presents a robustness analysis under increasing Gaussian noise. Across all four methods, the original examples consistently maintain a higher validity under noise than their counterfactuals. TSEVO and NG-CF show sharp drops in counterfactual validity (below 0.4 at $\epsilon = 0.4$), while their originals remain above 0.65 even at $\epsilon = 1.2$. COMTE shows a gradual decline, though originals still outperform counterfactuals. AB-CF CFEs degrade rapidly and the originals again show greater resilience. The confidence curves further confirm this robustness gap between the originals and CFEs.

Figure~\ref{fig:ecg200-dif} reveals another concern: The counterfactual robustness varies depending on the starting class of the user. For all methods, confidence differences between original and counterfactual instances show consistent asymmetries when split by recourse direction (class 0 to 1 vs.\ 1 to 0). This implies that the user experience with CFEs is not uniform; some users receive more robust and actionable recourse than others. This asymmetry raises concerns about the consistency of the model and CFEs and highlights the need for evaluation protocols that account for such directional biases.



\subsection{Alternative views}

This raises a critical question: are CFEs even suitable for achieving algorithmic recourse in the first place? It may be more appropriate to reconsider the goal of recourse from the perspective of practical utility, rather than theoretical optimality or elegance. While many CFE methods have been proposed, their real-world applicability remains underexplored. CFEs may be useful in certain scenarios but should not be treated as a universal solution for recourse. Recent work has shown that when unknown attributes of the user are introduced, CFEs may offer limited benefit compared to simpler explanations like reason codes~\cite{upadhyay_counterfactual_2025}. This highlights the importance of evaluating interpretability methods from a user-centered perspective, rather than relying solely on model-centric metrics.

To better formalize the distinction between explanations and actionable recommendations, Karimi et al.~\cite{karimi2022survey} differentiate between input-based CEs and recourse through feasible actions (CR). While CEs are changes in the input space defined as $\mathbf{x}^{CE} = \mathbf{x} + \boldsymbol{\delta}$ using a distance function like $dist(\mathbf{x}, \mathbf{x}^{CE})$, CRs are based on interventions in a set of feasible actions $\mathcal{A}(\mathbf{x})$:

\begin{equation}\label{eq:CFE_CR}
    \underset{\mathbf{a} \in \mathcal{A}(\mathbf{x})}{\arg\min} \; cost(\mathbf{x}, \mathbf{a}) \quad \text{s.t.} \quad f(\mathbf{x}) \neq f(\mathbf{x}^{CE}).
\end{equation}

\noindent Here, actions $\mathbf{a}$ are modeled as interventions in a structural causal model, where modifying one variable may affect others. Thus, the cost of changing the input vector ($cost(\mathbf{x}, \mathbf{x}')$) can differ significantly from the cost of taking actions in the real world ($cost(\mathbf{x}, \mathbf{a})$). In causal settings, the counterfactual instance becomes \emph{structural counterfactual} $\mathbf{x}^{SCF}(\mathbf{a}, \mathbf{x})$, where actions $\mathbf{a}$ are grounded in the causal graph.

However, defining actions through structural models requires access to causal knowledge, which may not always be available. This motivates alternative formulations that encode feasibility directly into the optimization objective. For example, Ates et al.~\cite{ates_counterfactual_2021} introduce a binary diagonal matrix $A$ in the cost function, where $A_{jj} = 1$ if feature $j$ is allowed to change. SG-CF~\cite{li_sg-cf_2022}, another optimization-based method for TSC, incorporates a shapelet-guided loss that balances validity, proximity, sparsity, and contiguity. These approaches embed constraints into $cost(\mathbf{x}, \mathbf{x}')$, enabling flexible but structured recourse generation. However, for multivariate time series, the change vector increases from $\boldsymbol{\delta} \in \mathbb{R}^d$ to $\mathbb{R}^{d \times m}$, raising the question whether the traditional formulations of $cost(\cdot, \cdot)$ remain meaningful.

Ultimately, whether by redefining the input space or integrating richer feasibility constraints, we argue that algorithmic recourse must be designed in collaboration with the end user. In this direction, Knijnenburg et al.~\cite{knijnenburg2015evaluating} proposed the User-Centric Evaluation Framework for recommender systems, highlighting six key dimensions: objective and subjective system aspects, user experience, interaction, personal characteristics, and situational context. Recently, Donoso et al.~\cite{donoso2023towards} extended this framework to the evaluation of explainable AI (XAI) systems. Following this line of reasoning, we contend that the development and evaluation of CFEs, especially for time series, should be tightly coupled with empirical studies of user experience to ensure that recourse methods are not only technically sound but also practically usable.

\section{Conclusion}

This position paper identifies two key limitations in current CFE methods for time series: the lack of user-centered design and a temporal blind spot in the evaluation. We show that strong performance on standard metrics does not imply practical utility. A method may be robust and valid, but it produces CFEs with high proximity or implausible suggestions that are unsuitable for end-users, such as patients. In addition, class-specific disparities in CFE performance reveal issues of fairness that could be overlooked. Current evaluations also ignore the sequential nature of time series in a real-world context. Local input changes may disrupt temporal coherence, leading to unrealistic or unfeasible recommendations. To be actionable, CFEs must reflect user context and respect temporal structure. We call for methods and evaluation frameworks that integrate both: moving beyond prediction flips to feasible, goal-directed interventions.

\bibliographystyle{splncs04}
\bibliography{reference3}



\newpage
\appendix

\section{More on experimental setup}\label{app:exp_setup}

\subsection{Classification model}

\begin{figure}
    \centering
    \includegraphics[width=0.95\linewidth]{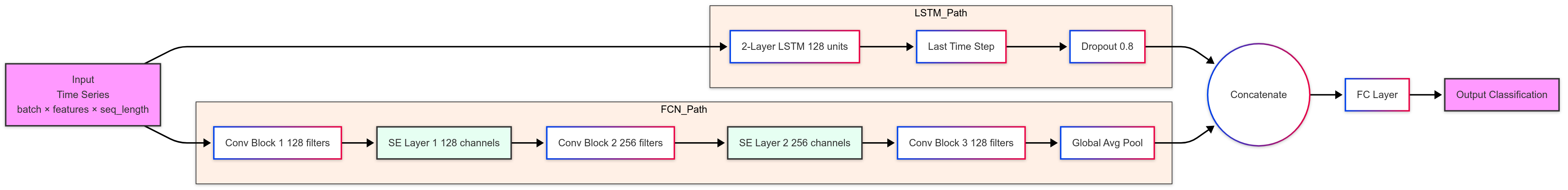}
    \caption{LSTM-FCN following \cite{karim_lstm_2018}.}
    \label{fig:lstmfcn}
\end{figure}

Given a training dataset $D$ of $N$ input-label pairs $\{(\mathbf{x}_i, y_i)\}_{i=1}^N$ with $\mathbf{x}_i$ a $d$-dimensional time series and $y_i$ a true class label, we train an LSTM-FCN classifier $f$ following \cite{karim_lstm_2018}. LSTM-FCN is a hybrid model that combines Long Short-Term Memory (LSTM) networks, which are effective at modelling temporal dependencies in sequential data, with Fully Convolutional Networks (FCNs), which capture local and hierarchical patterns through convolutional layers. This architecture has shown competitive results in TSC tasks due to its ability to capture both sequential and spatial features \cite{karim_lstm_2018}. The LSTM-FCN implemented in this study is shown in Figure \ref{fig:lstmfcn}, with parameter settings.

\subsection{Evaluation metrics}


Given a dataset of $N$ input-label pairs $\{(\mathbf{x}^{(i)}, y^{(i)})\}_{i=1}^N$, where $y^{(i)}$ is the true class label, we define the following metrics:

\noindent\textbf{Average Confidence}: The confidence of prediction for sample $i$ is defined as the probability assigned to the predicted class:
\[
{Conf}^{(i)} = f_{y^{(i)}}(\mathbf{x}^{(i)}), \quad \text{where } y^{(i)} = \arg\max_{c \in [1,k]} f_c(\mathbf{x}^{(i)}).
\]

\noindent Then, the average confidence across all samples is:
\[
\text{Average Confidence} (avgConf) = \frac{1}{N} \sum_{i=1}^{N} f_{y^{(i)}}(\mathbf{x}^{(i)}).
\]

\noindent\textbf{Average Validity}: The validity of a prediction is defined as 1 if the predicted class matches the true label, and 0 otherwise. When $\mathbf{x}$ is altered by noise, the validity is defined as 1 if the new predicted class matches the previous prediction.
\[
\text{Val}^{(i)} = 
\begin{cases}
1 & \text{if } y^{(i)} = \tilde{y}^{(i)} \\
0 & \text{otherwise}
\end{cases}
= \mathds{1}\left(y^{(i)} = \tilde{y}^{(i)}\right), \text{where $\tilde{y}$ is the predicted class}
\]

\noindent Then, the average validity is:
\[
\text{Average Validity} (avgVal)= \frac{1}{N} \sum_{i=1}^{N} \mathds{1}\left(y^{(i)} = \tilde{y}^{(i)}\right).
\]

\subsection{Other evaluation metrics}
Sparsity ($\ell_0$): This implementation measures sparsity between a time series $x$ and its counterfactual $\mathbf{x}$ at two levels: point-wise, using \texttt{np.isclose()} to compare individual values within a small tolerance, and segment-wise, by dividing the series into 10\% segments, averaging each, and comparing those means. Both return a score between 0 and 1 via \texttt{np.mean()}, with higher values indicating greater similarity.
\begin{equation}
    \ell_0 = \frac{1}{m} \sum_{h=1}^{m} \mathds{1} \left( |\mathbf{x}_i - \mathbf{x}'_i| \leq e \right)
\end{equation}
\text{where } $e$ = \texttt{1e-3}

\textbf{\(\ell_1\)-norm}: (Manhattan) distance: Calculates the sum of absolute differences between corresponding points in the original time series $\mathbf{x}$ and counterfactual $\mathbf{x}'$.
\begin{equation}
    \ell_1(\mathbf{x}, \mathbf{x}') = \sum_{i=1}^{m} |\mathbf{x}_i - \mathbf{x}'_i|
\end{equation}

\textbf{\(\ell_2\)-norm}: (Euclidean) distance: Computes the square root of the sum of squared differences between $\mathbf{x}$ and $\mathbf{x}'$, representing the straight-line distance. It emphasizes larger deviations.
\begin{equation}
    \ell_2(\mathbf{x}, \mathbf{x}') = \sqrt{\sum_{i=1}^{m} (\mathbf{x}_i - \mathbf{x}'_i)^2}
\end{equation}

\textbf{DTW}: Measures similarity between sequences that may vary in speed/time by finding the optimal alignment between the points in $\mathbf{x}$ and $\mathbf{x}'$.
\begin{equation}
    d_{\text{DTW}}(\mathbf{x}, \mathbf{x}') = \min_{\pi} \sqrt{\sum_{(i,j) \in \pi} (\mathbf{x}_i - \mathbf{x}'_j)^2}
\end{equation}
\noindent where \( \pi \) is the optimal alignment path minimizing the total cost.

\section{More on experimental results}\label{app:B}

\subsection{Detailed results}\label{sec:app_results}

This section presents the results underlying Figures \ref{fig:ecg200-dif} and \ref{fig:ecg200-proxspars} in Tables \ref{tab:ecg200-dif} and \ref{tab:ecg200-proxspars}.

\begin{table}[ht]
\centering
\caption{Table underlying Figure \ref{fig:ecg200-dif}. The table contains the mean and standard deviation of the distance in confidence between CFE and original instance (left), CFE with and without $\epsilon=0.2$ Gaussian noise (center), and original instance with and without $\epsilon=0.2$ Gaussian noise (right), for $n_{\text{valid}}$ CFEs generated with 4 CFE methods (x-axis) on the ECG200 dataset. Results are displayed separately for CFEs generated for class 1 (rather than 0, in orange) and class 0 (rather than 1, in purple).}
\label{tab:ecg200-dif}
\begin{tabular}{lcccccccc}
\toprule
& & & \multicolumn{2}{c}{CFE minus Orig.} & \multicolumn{2}{c}{CFE w/wo noise} & \multicolumn{2}{c}{Orig. w/wo noise} \\
 \cmidrule(l){4-5} \cmidrule(l){6-7} \cmidrule(l){8-9} \\
  \textbf{Method} & to-class & $n_{\text{valid}}$ & mean & sd & mean & sd & mean & sd \\ 
  \midrule
\multirow{2}{*}{AB-CF} &   1 &  24 & -0.19 & 0.24 & -0.00 & 0.08 & -0.09 & 0.18 \\ 
&   0 &  36 & -0.19 & 0.13 & -0.02 & 0.06 & -0.10 & 0.18 \\ 
\multirow{2}{*}{CoMTE} &   1 &  28 & 0.05 & 0.18 & -0.01 & 0.05 & -0.00 & 0.06 \\ 
 &   0 &  62 & -0.01 & 0.13 & 0.01 & 0.04 & -0.04 & 0.08 \\ 
\multirow{2}{*}{NG-CF} &   1 &  26 & -0.22 & 0.24 & -0.05 & 0.13 & -0.00 & 0.14 \\ 
 &   0 &  46 & -0.29 & 0.17 & 0.01 & 0.05 & -0.13 & 0.19 \\ 
\multirow{2}{*}{TSEVO} &   1 &  28 & -0.37 & 0.14 & -0.04 & 0.10 & 0.08 & 0.19 \\ 
 &   0 &  62 & -0.45 & 0.11 & -0.00 & 0.03 & -0.11 & 0.16 \\ 
   \bottomrule
\end{tabular}
\end{table}

\begin{table}[ht]
\centering
\caption{Table underlying Figure \ref{fig:ecg200-proxspars}. The table contains the mean and standard deviation for 3 proximity metrics ($\ell_1$, $\ell_2$ and DTW) and 2 sparsity metrics ($\ell_0$ and segment-based sparsity), over $n_{\text{valid}}$ CFEs (see Table \ref{tab:cfes_valid}). Results are displayed for CFEs generated for class 1 (rather than 0) and class 0 (rather than 1) separately.}
\label{tab:ecg200-proxspars}
\begin{tabular}{lccccccccccc}
  \toprule
& & \multicolumn{2}{c}{$\ell_1$} & \multicolumn{2}{c}{$\ell_2$} & \multicolumn{2}{c}{DTW} & \multicolumn{2}{c}{Tp sparsity $\ell_0$} & \multicolumn{2}{c}{Segment sparsity} \\
 \cmidrule(l){3-4} \cmidrule(l){5-6} \cmidrule(l){7-8} \cmidrule(l){9-10} \cmidrule(l){11-12} \\
  \textbf{Method} & to-class & mean & sd & mean & sd & mean & sd & mean & sd & mean & sd\\ 
  \midrule
\multirow{2}{*}{AB\_CF} & 1 & 17.31 & 14.00 & 3.37 & 1.84 & 1.31 & 0.74 & 0.67 & 0.21 & 0.64 & 0.23 \\ 
& 0 & 38.19 & 20.29 & 5.86 & 2.75 & 1.74 & 0.69 & 0.31 & 0.21 & 0.26 & 0.23 \\ 
  \multirow{2}{*}{CoMTE} &   1 & 43.98 & 20.80 & 6.13 & 2.74 & 2.92 & 1.11 & 0.00 & 0.01 & 0.01 & 0.04 \\ 
&   0 & 40.72 & 6.79 & 5.45 & 0.87 & 2.86 & 0.90 & 0.00 & 0.01 & 0.01 & 0.02 \\ 
  \multirow{2}{*}{NG\_CF} &   1 & 16.86 & 17.04 & 3.50 & 2.41 & 1.75 & 1.39 & 0.64 & 0.26 & 0.56 & 0.30 \\ 
& 0 & 19.65 & 12.98 & 3.43 & 1.74 & 1.78 & 0.98 & 0.51 & 0.25 & 0.44 & 0.25 \\ 
  \multirow{2}{*}{TSEVO} &   1 & 9.50 & 8.57 & 2.91 & 1.92 & 1.80 & 1.03 & 0.85 & 0.08 & 0.68 & 0.14 \\ 
& 0 & 20.12 & 14.17 & 4.52 & 2.14 & 2.29 & 1.24 & 0.67 & 0.19 & 0.51 & 0.20 \\ 
   \hline
\end{tabular}
\end{table}

\subsection{Additional results}\label{sec:results_otherdatasets}

\begin{figure}
    \centering
    \includegraphics[width=0.95\linewidth]{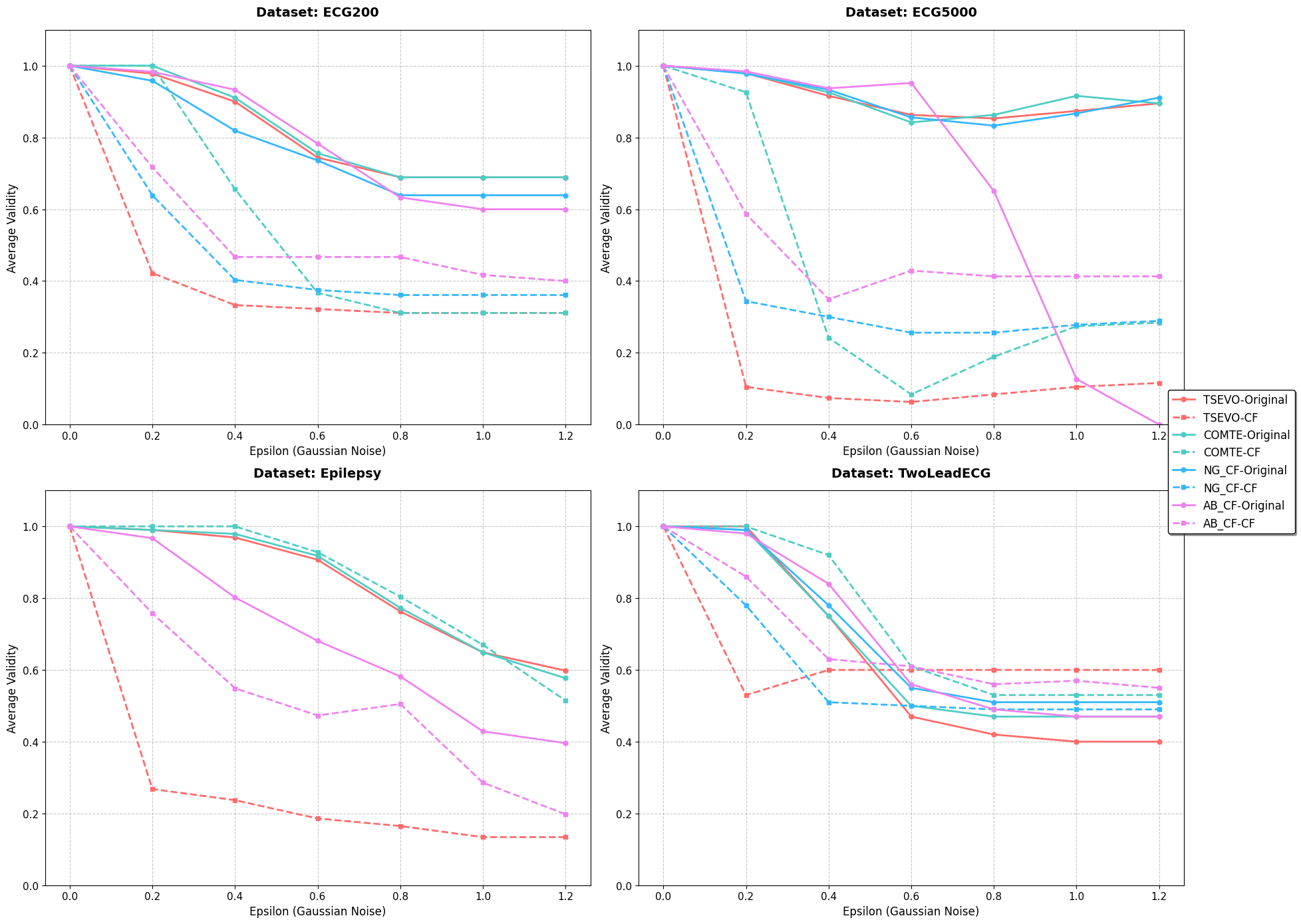}
    \caption{Robustness analysis of the CFE methods for all the datasets used}
    \label{fig:enter-label}
\end{figure}


\end{document}